\documentclass[a4paper]{article}
\usepackage{Odyssey2022}
\usepackage{epsfig,amssymb,amsmath}
\ninept

\usepackage[bottom]{footmisc}
\usepackage{multirow,booktabs,url}
\newcommand{\scell}[2][c]{%
  \begin{tabular}[#1]{@{}c@{}}#2\end{tabular}}

\usepackage[edges]{forest}

\definecolor{foldercolor}{RGB}{124,166,198}

\tikzset{pics/folder/.style={code={%
    \node[inner sep=0pt, minimum size=#1](-foldericon){};
    \node[folder style, inner sep=0pt, minimum width=0.3*#1, minimum height=0.6*#1, above right, xshift=0.05*#1] at (-foldericon.west){};
    \node[folder style, inner sep=0pt, minimum size=#1] at (-foldericon.center){};}
    },
    pics/folder/.default={20pt},
    folder style/.style={draw=foldercolor!80!black,top color=foldercolor!40,bottom color=foldercolor}
}

\forestset{is file/.style={edge path'/.expanded={%
    ([xshift=\forestregister{folder indent}]!u.parent anchor) |- (.child anchor)}, inner sep=1pt},
    this folder size/.style={edge path'/.expanded={%
    ([xshift=\forestregister{folder indent}]!u.parent anchor) |- (.child anchor) pic[solid]{folder=#1}}, inner xsep=0.75*#1},
    folder tree indent/.style={before computing xy={l=#1}},
    folder icons/.style={folder, this folder size=#1, folder tree indent=1.3*#1},
    folder icons/.default={12pt},
}
  
\title{A Study of Multimodal Person Verification Using Audio-Visual-Thermal Data}

\makeatletter
\def\name#1{\gdef\@name{#1\\}}
\makeatother
\name{{\em Madina Abdrakhmanova, Saniya Abushakimova, Yerbolat Khassanov, and Huseyin Atakan Varol}}

\address{Institute of Smart Systems and Artificial Intelligence,\\Nazarbayev University, Nur-Sultan City, Kazakhstan\\
{\small \tt \{madina.abdrakhmanova,saniya.abushakimova,yerbolat.khassanov,ahvarol\}@nu.edu.kz} }
\begin{document}
\maketitle

\begin{abstract}
In this paper, we study an approach to multimodal person verification using audio, visual, and thermal modalities.
The combination of audio and visual modalities has already been shown to be effective for robust person verification.
From this perspective, we investigate the impact of further increasing the number of modalities by adding thermal images.
In particular, we implemented unimodal, bimodal, and trimodal verification systems using state-of-the-art deep learning architectures and compared their performance under clean and noisy conditions.
We also compared two popular fusion approaches based on simple score averaging and the soft attention mechanism.
The experiment conducted on the SpeakingFaces dataset demonstrates the superior performance of the trimodal verification system.
Specifically, on the \textit{easy} test set, the trimodal system outperforms the best unimodal and bimodal systems by over 50\% and 18\% relative equal error rates, respectively, under both the clean and noisy conditions.
On the \textit{hard} test set, the trimodal system outperforms the best unimodal and bimodal systems by over 40\% and 13\% relative equal error rates, respectively, under both the clean and noisy conditions.
To enable reproducibility of the experiment and facilitate research into multimodal person verification, we made our code, pretrained models, and preprocessed dataset freely available in our GitHub repository\footnote{\url{https://github.com/IS2AI/trimodal_person_verification}\label{ft:github}}.
\end{abstract}

\section{Introduction}
\label{sec:intro}
Person verification is the general task of verifying a person's identity using various biometric characteristics.
The two widely used biometric features are voice (i.e., audio) and face (i.e., visual), corresponding to the speaker~\cite{DBLP:journals/spm/HansenH15} and the face~\cite{DBLP:conf/sibgrapi/Masi0HN18} verification tasks, respectively.
Recently, it has been observed that the combination of audio and visual modalities positively impacts the person verification task, yielding more accurate and robust systems~\cite{shon2019noise,qian2021audio}.
Presumably, increasing the number of modalities with the use of thermal images might boost the performance of a person verification system.

To date, most of the existing works on person verification have focused on unimodal approaches utilizing various biometric features, such as voice, face, fingerprints, iris, gait, and so on~\cite{minaee2019biometrics}.
Remarkable progress has been achieved in the use of voice~\cite{dehak2010front,DBLP:conf/interspeech/SnyderGPK17,nagrani2020voxceleb} and face~\cite{DBLP:conf/bmvc/ParkhiVZ15,DBLP:conf/cvpr/SchroffKP15} biometrics.
However, the performance of unimodal verification systems degrades substantially under challenging conditions~\cite{shon2019noise,qian2021audio}.
In response, several works have turned to multimodal approaches that introduce effective audio-visual fusion strategies.
For example,~\cite{DBLP:conf/icassp/SellDSEG18,alam2020analysis} combined the scores generated by unimodal verification systems.
In contrast,~\cite{shon2019noise,qian2021audio} explored fusion strategies using attention-based deep neural networks~\cite{DBLP:journals/corr/BahdanauCB14} and found that the proposed multimodal verification systems outperformed their unimodal counterparts and were more robust to noise.
 
Recently, consumer electronics manufacturers have started to equip smartphones with thermal cameras~\cite{one_pro,s62}, bringing to market mobile devices that incorporate the triplet of audio, visual, and thermal modalities.
A thermal camera presents new opportunities for a wide range of applications. 
In particular, it enables a more nuanced analysis of images in suboptimal physical environments (e.g., nighttime) and reduces the effects of variation in background, clothing, accessories, and appearance (e.g., makeup and contact lenses).
Additionally, a high-resolution thermal camera can provide more granular association of temperature values with facial features.
Its combination with other modalities can potentially strengthen multimodal person verification systems, by providing complementary information and alleviating the respective drawbacks of other modalities. 
Thermal images have been successfully deployed for speech recognition~\cite{DBLP:journals/tce/AndersonFT13}, lip reading~\cite{saitoh2006lip}, and person recognition using body images~\cite{DBLP:journals/sensors/NguyenHKP17}. 

In this work, we investigate whether the addition of thermal images as a third modality can improve the performance (accuracy and robustness) of audio-visual person verification systems. 
To the best of our knowledge, this is the first attempt at audio-visual-thermal person verification.
In particular, we adapted the SpeakingFaces dataset~\cite{abdrakhmanova2021speakingfaces} to the trimodal person verification task and shared the preprocessed data with the community.
We also developed unimodal, bimodal (i.e., audio-visual) and trimodal (i.e., audio-visual-thermal) verification systems using state-of-the-art deep learning architectures based on ResNet34~\cite{DBLP:conf/cvpr/HeZRS16}.
Additionally, we compared two popular fusion strategies based on simple score averaging~\cite{DBLP:conf/avbpa/LuceyC03,DBLP:conf/icassp/SellDSEG18} and the attention mechanism~\cite{DBLP:journals/corr/BahdanauCB14}.

We prepared two evaluation sets denoted as \textit{easy} and \textit{hard}, where the latter consists only of same gender pairs, and evaluated all models under two data conditions: 
\begin{itemize}
    \item[1)] \textit{Clean:} The training, validation, and test sets contained no augmentations or noise.
    \item[2)] \textit{Noisy:} Each of the three sets contained $30\%$ noise.
\end{itemize}


Experimental results on the \textit{easy} test set show that under the clean and noisy conditions the trimodal verification system outperforms the best unimodal system by 56\% and 51\% relative equal error rates (EERs), respectively.
The improvements over the best bimodal system under both conditions are over 18\% relative EERs.
On the \textit{hard} test set, the relative EER improvements under the clean and noisy conditions by the trimodal system over the best unimodal system are 52\% and 42\%, respectively.
The improvements over the best bimodal system are 18\% and 12\% relative EERs under the clean and noisy conditions, respectively.


The main contributions of our work are as follows:
\begin{itemize}
    \item[1)] We developed a trimodal person verification system based on audio-visual-thermal streams using state-of-the-art neural network architectures, and compared it with unimodal and bimodal systems.
    \item[2)] We evaluated the performance of the trimodal system under the clean and noisy conditions, and assessed the efficacy of two different fusion strategies. 
    \item[3)] We prepared an audio-visual-thermal dataset to facilitate research on multimodal person verification in this direction.
    \item[4)] We made our code, pretrained models, and dataset freely available$^{\ref{ft:github}}$ to enable experiment reproducibility.
\end{itemize}

The rest of the paper is organized as follows:
Section~\ref{sec:related} briefly reviews related work on person verification.
Section~\ref{sec:dataset} details the specifications of the SpeakingFaces dataset used in our experiments.
Section~\ref{sec:method} describes the architecture of the developed person verification systems.
Section~\ref{sec:exp_setup} presents the experimental setup. Section~\ref{sec:exp_results} discusses the obtained results. Section~\ref{sec:conclusion} concludes the paper and  highlights directions for future work.

\section{Related work}
\label{sec:related}
With the increasing popularity of biometric authentication in personal devices (e.g., mobile phones, tablets, computers, etc.), and in private entities (e.g., vehicles, houses, offices, etc.), robust and reliable person verification has become of paramount importance~\cite{bhattacharyya2009biometric}.
In recent years, many effective verification approaches have been proposed where promising results have been achieved~\cite{DBLP:journals/access/RuiY19}.
These approaches utilize various biometric features, including voice, face, fingerprint, palmprint, iris, and gait, to name a few.
Among the aforementioned biometrics, voice and face recognition has attracted substantial research interest in the academic community.

\subsection{Speaker verification}
Speaker verification aims to verify a person’s claimed identity using voice biometrics.
With the growing popularity of deep neural network (DNN), the leading modeling approach for speaker verification has transferred from the traditional Gaussian Mixture Model-Universal Background Model (GMM-UBM)~\cite{reynolds2000speaker} and \textit{i}-vector~\cite{dehak2010front} to deep speaker embedding representation learning.
A typical speaker embedding learning approach is \textit{d}-vector~\cite{variani2014deep}, where a fully-connected DNN is used to extract frame-level deep features.
These features are then averaged to form an utterance-level speaker representation.
Another popular speaker embedding learning approach is the time-delay neural network (TDNN) based \textit{x}-vector~\cite{DBLP:conf/interspeech/SnyderGPK17}, which has been shown to achieve state-of-the-art results in multiple datasets~\cite{snyder2018x}.
Recently, more advanced neural architectures, such as ResNet34, have been shown to further improve performance by extracting \textit{r}-vectors~\cite{zeinali2019but}.
In this work, we also employed ResNet34 to learn speaker embedding vectors from the audio stream.

\subsection{Face verification}
Face verification aims to verify a person’s claimed identity using visual images.
Since the impressive success of the AlexNet~\cite{DBLP:conf/nips/KrizhevskySH12} architecture in the ImageNet competition, deep convolutional neural network (DCNN)-based approaches have become the mainstream for the face verification task.
For example, DeepFace~\cite{taigman2014deepface} achieved human-level performance on the challenging LFW benchmark~\cite{LFWTech} for the first time.
Furthermore, in recent years, researchers have explored many different DCNN-based architectures for the face verification task, where remarkable improvements have been achieved.
These architectures include, but not limited to, DeepID~\cite{sun2015deepid3}, VGGNet~\cite{DBLP:journals/corr/SimonyanZ14a}, FaceNET~\cite{DBLP:conf/cvpr/SchroffKP15}, and ResFace~\cite{DBLP:conf/cvpr/HeZRS16}.
In our work, we will employ ResNet34 to extract embedding vectors from both visual and thermal images.


\subsection{Audio-visual person verification}
Although speaker verification and face verification have achieved remarkable progress in recent years, their performance decreases significantly under more challenging conditions.
For example, speaker verification is sensitive to changes in acoustic characteristics due to background noise, the person's mood (e.g., happy, angry, etc.), the uttered text, the recording device, distance, and so on.
Similarly, the performance of face verification is vulnerable to illumination, pose, emotion, distance.

To alleviate the drawbacks of the two verification approaches, researchers are turning to audio-visual verification approaches utilizing both modalities.
The initial work in this direction employed score-level fusion strategies~\cite{DBLP:conf/icassp/SellDSEG18,alam2020analysis,DBLP:conf/clear/LuqueMGAFMMMVH06}, where scores obtained from separately trained unimodal models were combined.
More recent works exploit attention-based fusion~\cite{shon2019noise,qian2021audio} to judiciously combine salient features from input modalities.
Overall, compared to unimodal verification systems, multimodal systems have been shown to be more accurate and robust, especially under noisy conditions.

To facilitate research on audio-visual person verification, various multimedia datasets have been introduced, such as VAST~\cite{DBLP:conf/lrec/TraceyS18}, JANUS~\cite{DBLP:conf/icassp/SellDSEG18}, VoxCeleb 1~\cite{nagrani2017voxceleb} \& 2~\cite{chung2018voxceleb2}.
Additionally, several person verification challenges have been organized~\cite{DBLP:conf/clear/StiefelhagenBBGMS06,sadjadi20202019}.
For example, the 2019 NIST speaker recognition evaluation (SRE) challenge~\cite{sadjadi20202019} proposed to use visual information in conjunction with audio recordings to increase the robustness of verification systems.
As a result, audio-visual fusion was found to result in remarkable performance gains (more than 85\% relative) over audio-only or face-only systems.
In this paper, we will investigate further increase of modalities by supplementing an audio-visual system with thermal images.

\section{Audio-Visual-Thermal Dataset}
\label{sec:dataset}
In this work, we utilized the SpeakingFaces dataset~\cite{abdrakhmanova2021speakingfaces} to train, validate, and test the person verification systems.
SpeakingFaces is a publicly available multimodal dataset comprised of audio, visual, and thermal data streams.
The dataset consists of 13,711 recordings of imperative sentences\footnote{Verbal commands given to virtual assistants and other smart devices, such as `turn off the lights', `play the next song', etc.} spoken by 142 speakers from different backgrounds (i.e., around 100 utterances per speaker).
The audio and visual streams were recorded using a webcam (Logitech C920 Pro HD), while the thermal stream was captured with a high-resolution thermal camera (FLIR T540).
The data were acquired approximately one meter away from the person.

\begin{figure}[t]
  \centering
  \includegraphics[width=1.0\columnwidth,trim={0.15cm 10.0cm 16.5cm 0.0cm},clip=true]{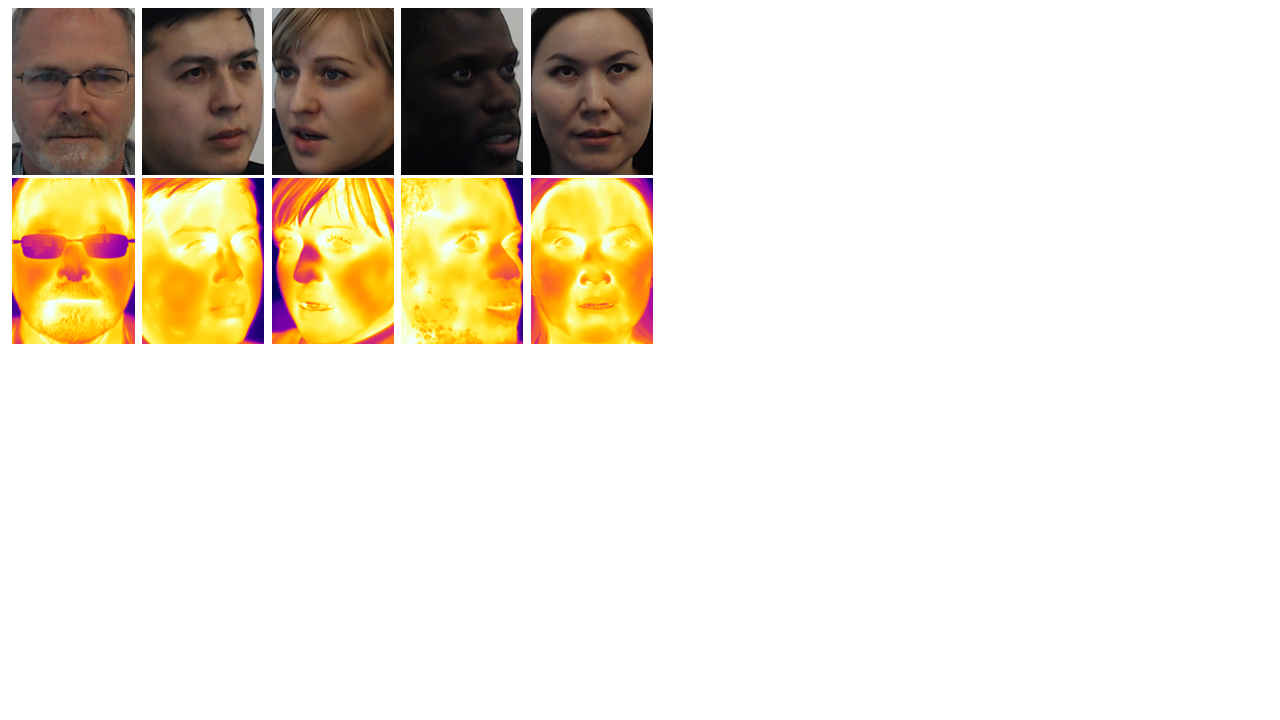}
  \vspace{-0.7cm}
  \caption{The samples of visual and thermal facial image pairs in the preprocessed SpeakingFaces dataset}
  \label{fig:faces}
\end{figure}

To prepare the dataset for the person verification task, we first detected bounding boxes of facial regions in visual images using RetinaFace~\cite{deng2019retinaface}.
Next, we mapped the detected bounding boxes onto the thermal images by aligning the visual and thermal streams.
Additionally, we manually filtered out from the final dataset those utterances where faces were not detected or where facial features were obscured based on previously documented artifacts.
This left 13,036 recordings in the final version.
We also downsampled the audio recordings to 16 kHz.
The samples of visual and thermal facial image pairs are shown in Figure~\ref{fig:faces} and the specifications of the preprocessed dataset are listed in Table~\ref{tab:dataset}.

\begin{table}[t]
\vspace{-0.2cm}
\caption{Preprocessed SpeakingFaces dataset specification}
\label{tab:dataset}
\vspace{0.2cm}
\centering
\renewcommand\arraystretch{1.1}
\setlength{\tabcolsep}{2.75mm}
    \begin{tabular}{l|c|c|c|c|c}
        \toprule
        \multicolumn{2}{l|}{\textbf{Category}}                      & \textbf{Train}& \textbf{Valid}& \textbf{Test} & \textbf{Total} \\
        \midrule
        \multicolumn{2}{l|}{\# Speakers}                            & 100           & 20            & 22            & 142 \\
        \multicolumn{2}{l|}{\# Utterances}                          & 9,307         & 1,843         & 1,886         & 13,036 \\
        \multicolumn{2}{l|}{Total duration}             & 10.4 hr       & 2.2 hr        & 2.3 hr        & 14.9 hr \\
        \bottomrule
    \end{tabular}
\vspace{-0.3cm}
\end{table}

\begin{figure}[t]
    \centering 
    \scriptsize
    \begin{forest}
        for tree={
            font=\sffamily,
            grow'=0,
            inner sep=1.75pt,
            folder indent=0.0em,
            folder icons,
            edge=densely dotted
        }
        [SpeakingFaces, this folder size=-5pt
            [data, this folder size=8pt
                [sub\_1, this folder size=8pt
                    [rgb, this folder size=8pt
                        [\textit{subID\_trialID\_2\_posID\_commandID\_frameID\_7.png}, is file]
                        [\textbf{\dots}, is file]]
                    [thr, this folder size=8pt
                        [\textit{subID\_trialID\_2\_posID\_commandID\_frameID\_6.png}, is file]
                        [\textbf{\dots}, is file]]
                    [wav, this folder size=8pt
                        [\textit{subID\_trialID\_2\_posID\_commandID\_1.wav}, is file]
                        [\textbf{\dots}, is file]]]
                [\textbf{\dots}, this folder size=8pt]
                [sub\_142, this folder size=8pt
                    [\textbf{\dots}, is file]]]
            [metadata, this folder size=8pt
                [\textit{train\_list.txt}, is file]
                [\textit{valid\_pairs\_easy.txt}, is file]
                [\textit{test\_pairs\_easy.txt}, is file]
                [\textit{valid\_pairs\_hard.txt}, is file]
                [\textit{test\_pairs\_hard.txt}, is file]
                ]
        ]
    \end{forest}
    \caption{The file structure of the SpeakingFaces dataset for the multimodal person verification task}\label{fig:file_struc}
\end{figure}
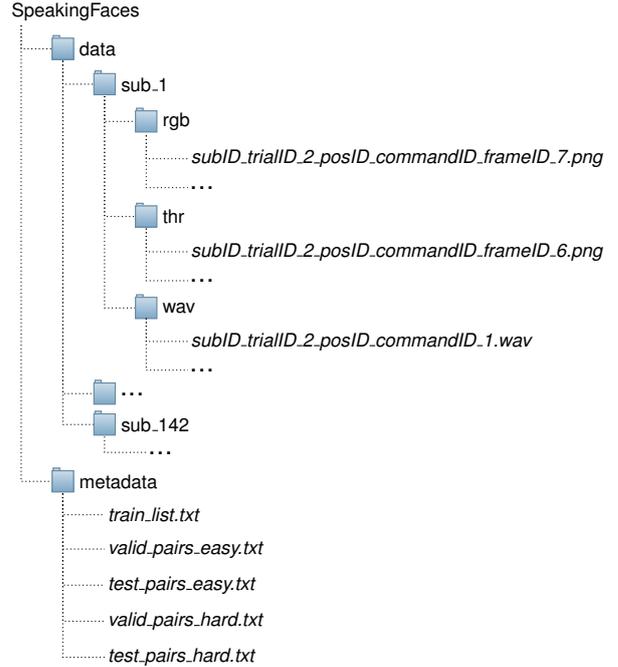

We prepared the train, validation, and test sets following the format of VoxCeleb~\cite{nagrani2017voxceleb}.
In particular, the train list contains the paths to the recordings and the corresponding subject identifiers. 
We designed \textit{easy} and \textit{hard} versions of the test and validation lists.
In the \textit{easy} version, the evaluation lists consist of randomly generated positive and negative pairs, with the numbers of pairs in the validation and test sets equal to 38,000 and 46,200, respectively.
In the \textit{hard} version, the negative pairs were randomly selected from subjects belonging to the same gender, resulting in 36,400 and 44,000 pairs in the validation and test sets, respectively.
Overall, the sets are gender balanced and have similar age and race distributions.
The file structure of the prepared SpeakingFaces dataset for multimodal person verification task is shown in Figure~\ref{fig:file_struc}.

To imitate more challenging scenarios, we deployed a noisy version of the train, validation, and test sets.
The noisy data were constructed using Algorithm 1 presented in~\cite{qian2021audio}.
In particular, for the visual stream, we applied vertical and horizontal motion blur, to imitate the motion of a person in front of the camera, and the Gaussian blur, to imitate other noises.
For the thermal stream, we applied only the Gaussian blur.
For the audio stream, we added three kinds of noises from the Musan dataset~\cite{DBLP:journals/corr/SnyderCP15}. 
In addition, for all streams, we used random noise sampled from the standard normal distribution (i.e., $\mathcal{N}(0,1)$) to imitate a broken data stream.

\section{System Architecture}
\label{sec:method}
\begin{figure}[thb]
  \centering
  \includegraphics[width=0.65\columnwidth,trim={0.0cm 8.75cm 22.5cm 0.0cm},clip=true]{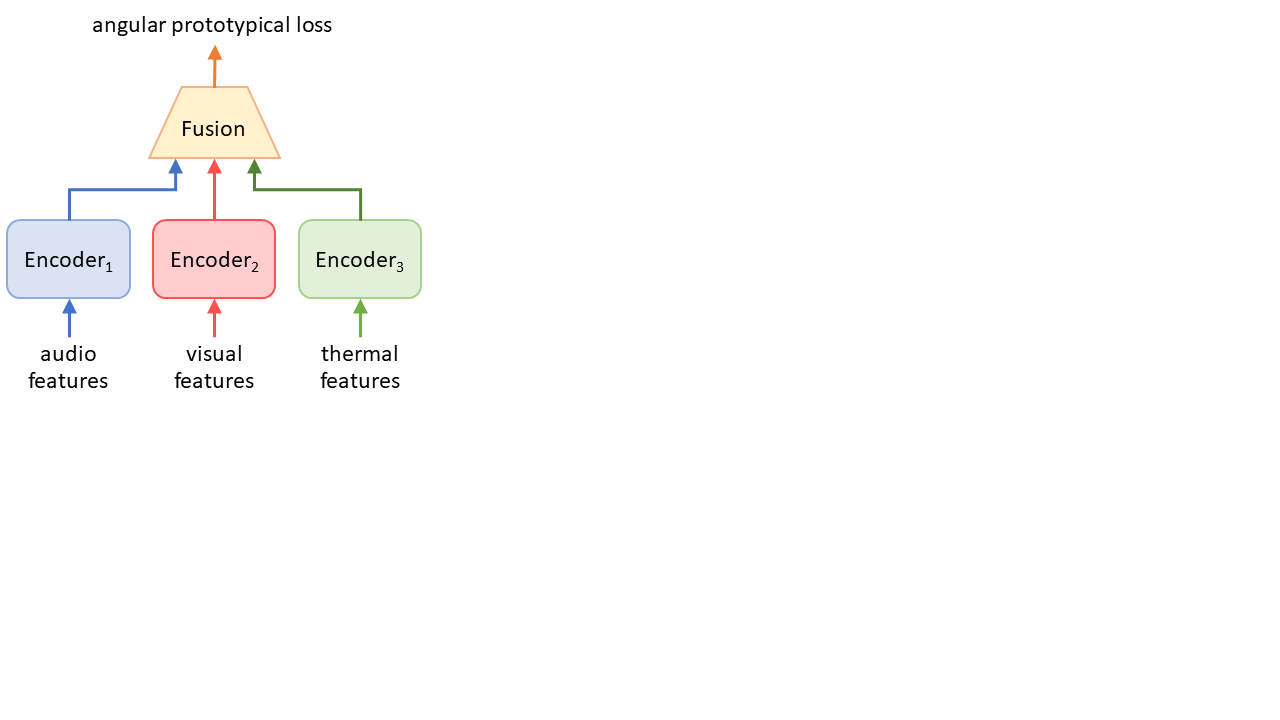}
  \caption{The audio-visual-thermal person verification network architecture}
  \label{fig:arch}
\end{figure}

We followed~\cite{chung2020defence} to implement the multimodal person verification system. 
The system architecture is depicted in Figure~\ref{fig:arch} which consists of two main modules: encoder and fusion.
The encoder module is responsible for extracting high-level feature representations from different modalities, and the fusion module is used to combine the extracted features.

\subsection{Encoder module}
The encoder module $Encoder(\cdot)$ is based on the ResNet34~\cite{DBLP:conf/cvpr/HeZRS16} network.
Specifically, for image input (i.e., visual and thermal), we used a variation of ResNet34, in which the number of channels in each residual block is halved in order to reduce computational costs.
For audio, we used another variation of ResNet34, in which average-pooling was replaced with self-attentive pooling (SAP)~\cite{DBLP:conf/odyssey/CaiCL18} to aggregate frame-level features into an utterance-level representation.

The encoder takes in a raw feature $x_{i}$ and outputs the corresponding embedding vector representation $e_{i}\in\mathbb{R}^{512}$:
\begin{equation}
e_{i} = Encoder(x_{i})
\end{equation}
where $i\in\{a,v,t\}$ is used to represent the stream source (i.e., audio, visual, and thermal).
We trained a separate encoder module for each data stream.

\subsection{Fusion module}
As a fusion module, we tried two different approaches: 1)~attention mechanism and 2)~score averaging.

\subsubsection{Attention mechanism}
In the attention-based approach, we implemented an embedding-level fusion similar to that in~\cite{shon2019noise}, where the fusion module can pay attention to a salient modality among audio $e_a$, visual $e_v$, and thermal $e_t$ representations.
Specifically, it first computes the attention score $\hat{\alpha}_{\{a,v,t\}}\in \mathbb{R}^3$ as follows:
\begin{equation}
\hat{\alpha}_{\{a,v,t\}} = \mathbf{W}[e_a,e_v,e_t]+\mathbf{b}
\end{equation}
where the weight matrix $\mathbf{W}\in\mathbb{R}^{3\times1536}$ and the bias vector $\mathbf{b}\in\mathbb{R}^3$ are learnable parameters.
Next, the fused person embedding vector $e_p$ is produced by the weighted sum:
\begin{equation}
e_p=\sum_{i\in\{a,v,t\}}\alpha_ie_i\text{\ \ , where\ \ } \alpha_i=\frac{\exp(\hat{\alpha}_i)}{\sum_{k\in\{a,v,t\}}\exp(\hat{\alpha}_k)}
\end{equation}
The attention-based fusion for the bimodal systems (i.e., audio-visual) was designed in a similar manner, using two modalities instead of three.
Note that the attention-based fusion module is jointly trained with the encoder modules in an end-to-end fashion.

\subsubsection{Score averaging}
In the score-level fusion, we simply averaged the scores $s_i\in[0,2]$ produced by the unimodal verification systems, where a score is the Euclidean distance between two normalized speaker embeddings.
For example, for the trimodal system, the final score $s_{\textit{final}}$ is computed as follows:
\begin{equation}
s_{\textit{final}} = \frac{\sum_{i\in\{a,v,t\}}s_i}{3}
\end{equation}
Likewise, for the bimodal system, the final score is computed by averaging the scores of two unimodal systems. 

\section{Experimental Setup}
\label{sec:exp_setup}
All the person verification models were trained on a single V100 GPU running on the NVIDIA DGX-2 server using the training set.
All hyper-parameters were tuned using the validation set, and the settings for each modality were tuned separately.
The best performing model on the validation set was evaluated using the test set.

At the training stage, input features from the audio stream were generated by randomly extracting a two-second segment from each recording and then transforming it into 40-dimensional mel spectogram features.
For the visual and thermal streams, we extracted the first frame of a recording. 
Model parameters were optimized using the angular prototypical~\cite{chung2020defence} loss function. 
Each batch contained all the 100 speakers of the train set with nine utterances per speaker.
As a regularization, we applied weight decay tuned for each model separately.
We trained each model three times with a different seed number and report the mean and standard deviation of the results.
The exact system implementation details are provided in our GitHub repository\textsuperscript{\ref{ft:github}}.

At the evaluation stage, from each test recording, we extracted ten two-second audio segments at regular intervals, and ten equidistantly spaced image frames.
In the multimodal setting with attention, for a given utterance, each of the ten audio segments was paired with an image frame from either or both modalities.
We computed a similarity score between all possible combinations (10×10 = 100) from each pair of utterances present in the evaluation sets.
As a similarity score, we used the Euclidean distance, and the mean of the 100 similarities was used as the final score.
The performance of the models was evaluated using the EER metric. 


\section{Experimental Results}
\label{sec:exp_results}
\begin{table*}[htb]
\caption{EER (\%) performance results (mean $\pm$ std) of person verification systems on the \textit{easy} evaluation set under two conditions:\\1) \textit{clean} and 2) \textit{noisy}. Under the \textit{noisy} condition, the percentages of noisy samples in the train, validation, and test sets were set to 30\%.}
\label{tab:results}
\vspace{2mm}
\centering
\renewcommand\arraystretch{1.1}
\setlength{\tabcolsep}{3.5mm}
    \begin{tabular}{l|c|cc|cc}
        \toprule
        \multirow{2}{*}{\textbf{Modality}}  & \multirow{2}{*}{\textbf{Fusion}}  &   \multicolumn{2}{c|}{\textit{\textbf{clean}}}             & \multicolumn{2}{c}{\textit{\textbf{noisy}}} \\\cline{3-6}
                                        &                           &  \textbf{valid}           & \textbf{test}         & \textbf{valid}    & \textbf{test} \\
        \midrule
        Audio                           & -                         & $10.82\pm0.39$            & $9.29\pm0.15$         & $15.03\pm0.54$    & $11.86\pm0.30$ \\
        Visual                          & -                         & $4.04\pm0.89$             & $4.09\pm0.86$         & $5.81\pm0.67$     & $4.36\pm0.93$ \\
        Thermal                         & -                         & $10.30\pm0.74$            & $10.58\pm1.28$        & $11.00\pm1.18$    & $12.72\pm0.93$ \\\hline
        \multirow{2}{*}{Audio-Visual}   & Attention                 & $4.42\pm0.85$             & $3.69\pm0.44$         & $5.46\pm1.01$     & $4.10\pm1.07$ \\.  
                                        & Score averaging           & $2.39\pm0.42$             & $2.20\pm0.58$         & $3.69\pm0.48$     & $2.61\pm0.45$ \\\cline{1-6}
        \multirow{2}{*}{Audio-Visual-Thermal}       & Attention                 & $3.44\pm0.60$             & $2.90\pm0.15$         & $4.02\pm1.02$     & $3.74\pm1.04$ \\
                                        & Score averaging           & \textbf{2.26 $\pm$ 0.10}  & \textbf{1.80 $\pm$ 0.31} & \textbf{2.92 $\pm$ 0.22} & \textbf{2.13 $\pm$ 0.47} \\
        \bottomrule
\end{tabular}
\end{table*}

\begin{table*}[htb]
\caption{EER (\%) performance results (mean $\pm$ std) of person verification systems on the \textit{hard} evaluation set under two conditions:\\1) \textit{clean} and 2) \textit{noisy}. Under the \textit{noisy} condition, the percentages of noisy samples in the train, validation, and test sets were set to 30\%.}
\label{tab:results_samegender}
\vspace{2mm}
\centering
\renewcommand\arraystretch{1.1}
\setlength{\tabcolsep}{3.5mm}
    \begin{tabular}{l|c|cc|cc}
        \toprule
        \multirow{2}{*}{\textbf{Modality}}  & \multirow{2}{*}{\textbf{Fusion}}  &   \multicolumn{2}{c|}{\textit{\textbf{clean}}}             & \multicolumn{2}{c}{\textit{\textbf{noisy}}} \\\cline{3-6}
                                        &                           &  \textbf{valid}           & \textbf{test}         & \textbf{valid}    & \textbf{test} \\
        \midrule
        Audio                           & -                         & $15.14\pm0.27$            & $14.13\pm0.12$         & $21.34\pm0.88$    & $17.99\pm1.01$ \\
        Visual                          & -                         & $5.60\pm0.91$             & $5.23\pm0.68$         & $6.85\pm0.86$     & $5.84\pm1.58$ \\
        Thermal                         & -                         & $13.21\pm1.07$            & $12.34\pm1.28$        & $13.25\pm1.32$    & $14.81\pm2.33$ \\\hline
        \multirow{2}{*}{Audio-Visual}   & Attention                 & $5.72\pm0.98$             & $4.89\pm0.42$         & $6.71\pm1.15$     & $5.61\pm1.61$ \\
                                        & Score averaging           & $3.48\pm0.64$             & $3.10\pm0.42$         & $5.33\pm0.43$     & $3.89\pm0.58$ \\\cline{1-6}
        \multirow{2}{*}{Audio-Visual-Thermal}       & Attention                 & $4.40\pm0.83$             & $4.72\pm1.49$         & $5.18\pm1.26$     & $5.18\pm1.73$ \\
                                        & Score averaging           & \textbf{3.07 $\pm$ 0.55}  & \textbf{2.53 $\pm$ 0.38} & \textbf{3.96 $\pm$ 0.12} & \textbf{3.40 $\pm$ 0.78} \\
        \bottomrule
\end{tabular}
\end{table*}

We evaluated the performance of unimodal and multimodal person verification systems on \textit{easy} and \textit{hard} evaluation lists under 
two data conditions: 1) \textit{clean} and 2) \textit{noisy}.
The noisy versions of the train, validation, and test sets were generated based on the procedure described in Section~\ref{sec:dataset}.

\begin{figure}[b]
    \centering
    \includegraphics[width=0.5\columnwidth,clip=true]{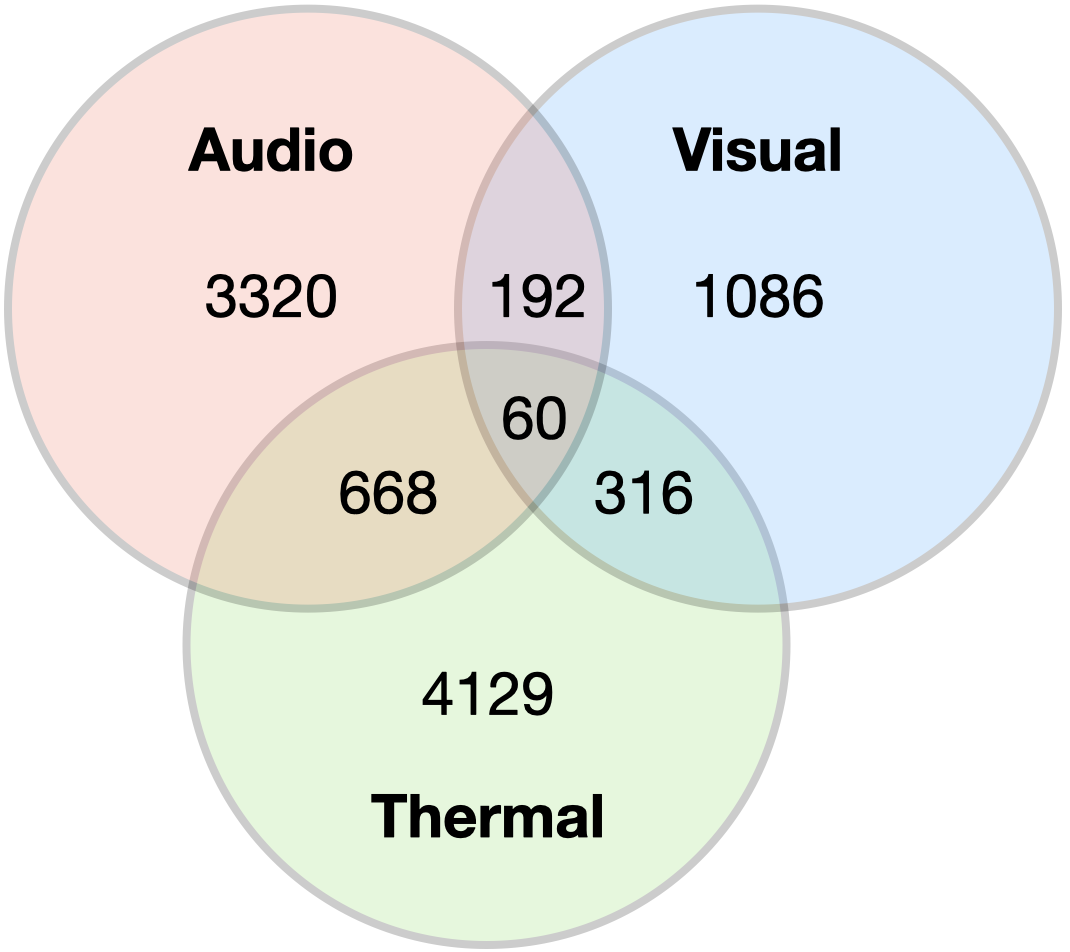}
    \caption{The verification error statistics of unimodal systems}
    \label{fig:ver_errors}
\end{figure}

\subsection{Unimodal Person Verification}
The results of the unimodal person verification experiment on the \textit{easy} test set are given in the first part of Table~\ref{tab:results}.
When train and evaluation sets are clean, the best EER performance is achieved by the visual modality (4.09\%), followed by the audio modality (9.29\%) and then the thermal modality (10.58\%).
Under the noisy condition, the performance degrades by 28\%, 7\%, and 20\% relative EER for the audio, visual, and thermal systems, respectively.
According to these results, the visual system performs the best and is more robust to corrupted data.

Interestingly, the results in Table~\ref{tab:results_samegender} show that the best EER performance on the \textit{hard} test set, under the clean condition, is achieved by the visual modality (5.23\%), followed by the thermal modality (12.34\%) and then the audio modality (14.13\%).
Under the noisy condition, the performance degrades by 27\%, 12\%, and 20\% relative EER for the audio, visual, and thermal systems, respectively.
These results indicate that the visual system remains superior, but the thermal modality outperforms the audio in differentiating subjects of the same gender\footnote{At least in our experimental settings.}.

To further examine the performance of the unimodal systems, we computed accuracy on the \textit{easy} test set under the \textit{clean} condition, separating the same- and opposite-gender pairs.
The results are presented in the first part of Table~\ref{tab:results_acc}.
Following the observations from the above, when a given pair of subjects belong to the same gender, the visual modality (95.07\%) is superior, while the thermal modality (88.06\%) is better than the audio modality (86.72\%). 
However, the audio modality (98.45\%) performs the best in distinguishing subjects of opposite gender.

We also analysed the verification errors made by the unimodal systems on the \textit{easy} test set under the \textit{clean} condition (see Figure~\ref{fig:ver_errors}).
We observed that the number of overlapping errors between different modalities is lower than the errors made by a single modality.
This indicates that these modalities posses strong complementary properties, and therefore, multimodal systems that can combine them effectively have good potential. 
Future work should focus on analysing these errors in great detail to identify the weaknesses and strengths of each modality.

\begin{table}[t]
\caption{Accuracy (\%) results (mean $\pm$ std) on the \textit{easy} test set. The data condition is \textit{clean} and multimodal systems are based on score fusion. Bimodal: Audio-Visual. Trimodal: Audio-Visual-Thermal.}
\label{tab:results_acc}
\vspace{2mm}
\centering
\renewcommand\arraystretch{1.1}
    \begin{tabular}{l|cc|c}
        \toprule
        \textbf{Modality}       & \textbf{\scell{Same\\gender}} & \textbf{\scell{Opposite\\gender}} & \textbf{Overall}\\
        \midrule
        Audio                   &   $86.72\pm1.88$              & $98.45\pm1.35$                    & $89.79\pm1.73$\\
        Visual                  &   $95.07\pm0.86$              & $98.28\pm0.90$                    & $95.91\pm0.87$\\
        Thermal                 &   $88.06\pm1.75$              & $93.24\pm0.70$                    & $89.41\pm1.29$\\\hline
        Bimodal                 &   $97.07\pm0.73$              & $99.87\pm0.17$                    & $97.80\pm0.58$\\\hline
        Trimodal                &   \textbf{97.61 $\pm$ 0.39}   & \textbf{99.87 $\pm$ 0.08}         & \textbf{98.20} $\pm$ \textbf{0.31}\\
        \bottomrule
\end{tabular}
\end{table}

\subsection{Multimodal Person Verification}

The experimental results for our bimodal (audio-visual) and trimodal (audio-visual-thermal) verification models are presented in the second and third parts of Table~\ref{tab:results} and Table~\ref{tab:results_samegender}, respectively.
The models were constructed using two fusion methods, soft attention and score averaging, as mentioned in the previous sections.
The latter approach provides superior performance for both bimodal and trimodal verification systems, which is consistent with the observations made in~\cite{qian2021audio}, but different from the findings in~\cite{shon2019noise}.

The experimental results show that the multimodal systems outperform the unimodal systems. 
For the \textit{easy} test set, the best bimodal system reduced EER by 46\% (from 4.09 to 2.20) and 40\% (from 4.36 to 2.61) relative to the visual system under the clean and noisy conditions, respectively.
The best trimodal system reduced EER by 56\% (from 4.09 to 1.80) and 51\% (from 4.36 to 2.13) under the clean and noisy conditions, respectively.
Remarkably, these improvements were achieved by simply averaging the scores of the unimodal systems.

For the \textit{hard} test set, we observed similar behaviour. The best bimodal system reduced EER by 41\% (from 5.23 to 3.10) and 33\% (from 5.84 to 3.89) relative to the visual system under the clean and noisy conditions, respectively.
The best trimodal system reduced EER by 52\% (from 5.23 to 2.53) and 42\% (from 5.84 to 3.40) under the clean and noisy conditions, respectively.

The trimodal system achieved better results than the bimodal system on both evaluation sets under the clean and noisy conditions.
On the \textit{easy} test set, EERs were improved by 18\% relative for both conditions (from 2.20 to 1.80, and from 2.61 to 2.13).
On the \textit{hard} test set, the trimodal system surpassed the bimodal system by 18\% (from 3.10 to 2.53) and 13\% (from 3.84 to 3.40) relative EERs, under the \textit{clean} and \textit{noisy} conditions respectively.

The analysis of the attention network parameters suggests that the mechanism learned to prioritize the visual stream when the streams were not corrupted.
In contrast, the network focused on all the three streams when the data were noisy. 
Therefore, augmenting the train set with noise plays an important role in learning more robust features, and future work should study different augmentation methods for the multimodal person verification task.
Unfortunately, combining three modalities using the soft attention mechanism is challenging due to the instability of the training process, and further studies should be conducted to explore other attention methods.
Overall, the results suggest that the addition of thermal image data does indeed enhance the EER performance of multimodal person verification systems.




The importance of supplementing the audio-visual modalities with the thermal images is also accentuated in Table~\ref{tab:results_acc}.
The bimodal system is able to take the best of the two streams and drastically boost the performance in handling both same- and opposite-gender pairs.
The addition of the thermal stream strengthens the performance of the multimodal system even further for the same gender pairs, but has an insignificant effect on the opposite gender pairs.

\section{Conclusion}
\label{sec:conclusion}
In this work, we explored multimodal learning for the person verification task using audio, visual and thermal data streams.
We adapted the SpeakingFaces dataset to build state-of-the-art deep learning-based models.
Specifically, we developed unimodal, bimodal, and trimodal verification systems and compared their performance on easy and hard evaluation lists under the clean and noisy data conditions.
Although the visual modality achieved the best result among the unimodal systems, the audio was superior in distinguishing subjects of opposite gender.
We also observed that the thermal modality performed better than the audio modality in differentiating between subjects of the same gender.
For the multimodal systems, we compared two popular fusion strategies based on simple score averaging and the soft attention mechanism.
The former achieved a substantial performance gain for the trimodal system compared to the other systems.
Specifically, the relative EER improvements over the visual-based system were 56\% and 51\%, and 18\% and 20\% over the audio-visual system, under the clean and noisy data conditions, respectively.

In the future, we plan to thoroughly analyze the errors made by the different unimodal systems to identify the weaknesses and strengths of each modality.
We also plan to implement more advanced attention mechanisms and study the impact of different data augmentation techniques.
The robustness evaluation of multimodal verification systems at different noise rates is also a topic for future work.

\bibliographystyle{IEEEbib}
\bibliography{main}

%

\end{document}